\pdfoutput=1

\documentclass[11pt]{article}

\usepackage{acl}

\usepackage{times}
\usepackage{latexsym}

\usepackage[T1]{fontenc}

\usepackage[utf8]{inputenc}

\usepackage{microtype}
\usepackage{graphicx}
\usepackage{caption}
\usepackage{subcaption}
\usepackage{amsmath}
\usepackage{amsfonts}

\title{A Dataset and BERT-based Models for Targeted Sentiment Analysis on Turkish Texts}

\author{M. Melih Mutlu \\
  Department of Computer Engineering \\
  Boğaziçi University \\
  \texttt{melih.mutlu@boun.edu.tr} \\\And
  Arzucan Özgür \\
  Department of Computer Engineering \\
  Boğaziçi University \\
  \texttt{arzucan.ozgur@boun.edu.tr} \\}

\begin{document}
\maketitle

\begin{abstract}

Targeted Sentiment Analysis aims to extract sentiment towards a particular target from a given text. It is a field that is attracting attention due to the increasing accessibility of the Internet, which leads people to generate an enormous amount of data. Sentiment analysis, which in general requires annotated data for training, is a well-researched area for widely studied languages such as English. For low-resource languages such as Turkish, there is a lack of such annotated data. We present an annotated Turkish dataset suitable for targeted sentiment analysis. We also propose BERT-based models with different architectures to accomplish the task of targeted sentiment analysis. The results demonstrate that the proposed models outperform the traditional sentiment analysis models for the targeted sentiment analysis task.


\end{abstract}

\section{Introduction}

\begin{table*}[!htbp]
\centering
\begin{tabular}{ p{10cm} p{2cm} p{2cm}  }
\hline
\textbf{Tweet} & \textbf{Sentence \newline Sentiment} & \textbf{Targeted \newline Sentiment} \\
\hline
\textit{coca cola} daha iyi lezzet olarak \newline (\textit{coca cola}'s taste is better)  & positive & positive \\
\textit{whatsapp} çöktü de biraz rahatladım bildirimlerden kurtuldum \newline (\textit{whatsapp} is crashed so I'm little relieved, got rid of notifications)  & positive & negative \\
\hline
\end{tabular}
\caption{
Sample tweets from the dataset. Targets are shown in italics. Sentences are annotated with respect to overall sentence sentiment and targeted sentiment which represent the sentiment towards the target. English translations are provided in parenthesis.
}
\label{tbl:dataset_ex}
\end{table*}

The increasing availability of the Internet and the growing number of online platforms allowed people to easily create online content. Because of the value of mining the people's opinions, the sentimental information contained in this online data makes sentiment analysis (SA) an interesting topic. It is an area that is attracting the attention not only of academic researchers, but also of businesses and governments \cite{survey} and has become a rapidly growing field, as evidenced by the number of recent SA papers published \cite{gov_org_comp}.

The problem with traditional sentiment analysis is that it cannot capture the different attitudes toward multiple aspects in a given text. For example, if the given text is \textit{``Phones from this brand are great, but I don't really like their laptops''}, the sentiment towards the two targets \textit{``phone''} and \textit{``laptop''} are positive and negative, respectively. Traditional sentiment analysis methods would not be able to detect this opposing sentiment for \textit{``phone''} and \textit{``laptop''}, but would assign an overall sentiment for the text. Targeted Sentiment Analysis (TSA) aims to overcome this challenge and extracts sentiment from a given text with respect to a specific target. One of the challenges of TSA is the lack of available datasets. Both TSA and SA require labeled datasets. Collecting data from various sources and labeling them, which is mostly done manually, is an expensive process. Although the number of datasets suitable for SA has recently increased due to new studies in the SA area, not all SA datasets are usable for TSA \cite{targetedsentimentanalysis}. TSA requires more refined datasets. The labels should reflect the sentiment toward targets rather than the overall sentiment of the sentences. 

English is the most studied language for sentiment analysis \cite{englishsa}. 
SA models that perform satisfactorily for English do not seem to always work with similar performance for Turkish \cite{kaya}. In this work, we create a manually annotated dataset from Twitter specifically labeled for both traditional and targeted sentiment analysis in Turkish. Then, we experiment with different model architectures for the Turkish TSA task. Experimental results demonstrate that our techniques outperform traditional sentiment analysis models.

\subsection{Problem Definition}

Let $E$ denotes all entities in a given document $D$ such that:
\begin{align*}
D &=\{w_1, \dots, w_k\} \text{ each } w \text{ is a word; } k \in \mathbb{Z^+} \\
E &= \{e_1, \dots, e_l\} \text{ each } e \text{ is an entity; } l \in \mathbb{Z^+} \\
T &=\{t_1, \dots, t_m\} \text{ } t_i \text{ is a target; } t_i\in E \text{; } m,i \in \mathbb{Z^+}
\end{align*}

The objective of targeted sentiment analysis is to find all sentiment $(s_i, t_i)$ pairs in document $D$ where $t_i$ is a target from $T$ and $s_i$ is the sentiment toward $t_i$.

\section{Related Work}

One of the challenges of targeted sentiment analysis is identifying contexts associated with target words in the sentiment classification. Early methods for understanding the relationship between the target and the rest of the sentence rely on hand-crafted feature extractions and rule-based techniques  \cite{ding2008holistic, jiang-etal-2011-target}. Recurrent neural networks (RNN) have been implemented for sentiment analysis in the recent years. It achieved improved results compared to earlier methods \cite{dong-etal-2014-adaptive, nguyen-shirai-2015-phrasernn, baktha2017investigation}. Two RNNs are used to obtain the context from both left and right and combine the context knowledge in \cite{tang2016effective}. Attention mechanisms are recently added into RNN-based methods to model the connection between each word and the target \cite{wang2016attention, ma2017interactive, zhang-etal-2020-target}.

\citet{vaswani2017attention} introduced the transformer architecture consisting of encoder and decoder blocks based on self-attention layers. Bidirectional Encoder Representations from Transformers (BERT) has been introduced and shown to achieve the state-of-the-art in various NLP tasks in \cite{devlin-etal-2019-bert}. BERT has recently become a widely used approach for sentiment analysis in many languages \cite{sun-etal-2019-utilizing, li-etal-2019-exploiting}. \citet{koksal2021twitter} provide a Twitter dataset in Turkish for sentiment analysis called BounTi. It consists of Twitter data which are about predefined universities and manually annotated by considering sentimental polarities towards these universities. They propose a BERT model fine-tuned using the BounTi dataset to identify sentiment in Turkish tweets.

\section{Dataset}

\begin{figure*}
     \centering
     \begin{subfigure}[t]{0.3\textwidth}
         \centering
         \includegraphics[width=\textwidth]{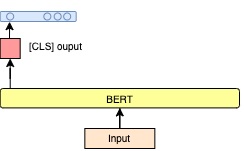}
         \caption{[CLS] output}
         \label{fig:cls}
     \end{subfigure}
     \hfill
     \begin{subfigure}[t]{0.3\textwidth}
         \centering
         \includegraphics[width=\textwidth]{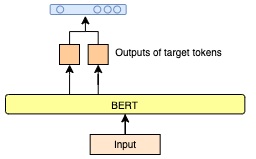}
         \caption{Max-pooling outputs of target tokens}
         \label{fig:maxpool}
     \end{subfigure}
     \hfill
     \begin{subfigure}[t]{0.3\textwidth}
         \centering
         \includegraphics[width=\textwidth]{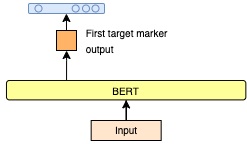}
         \caption{The output of the first target marker}
         \label{fig:marker}
     \end{subfigure}
        \caption{An overview of architectures to get and handle outputs from BERT}
        \label{fig:output}
\end{figure*}

Twitter is a commonly used source of sentiment classification dataset in the literature \cite{jiang-etal-2011-target, severyn-moschitti-2015-unitn, kruspe-etal-2020-cross}. In this study, we also create a Twitter dataset with 3952 tweets whose timestamps span a six-month period between January 2020 and June 2020. The tweets are collected via the official Twitter API by separately searching our 6 targets selected from famous companies and brands.

This dataset is manually annotated with three labels, positive, negative, and neutral. Two factors are considered in the annotation process, namely sentence sentiment and targeted sentiment. Each tweet has the following two labels. The sentence sentiment label expresses the overall sentiment of the sentence, regardless of the target word, as in traditional sentiment analysis techniques. On the other hand, the targeted sentiment label reflects the sentiment for the target in that sentence. The collected tweets are annotated separately by two annotators (one of the authors and a volunteer annotator) who are native Turkish speakers. 
Cohen's $\kappa$ \cite{cohen} is used to demonstrate inter-annotator agreement and is calculated as 0.855. In case of conflict between annotators, they re-evaluated the conflicting tweets. After re-evaluation, tweets on which the annotators agree  are retained and conflicting tweets are removed from the dataset.

Table \ref{tbl:dataset_ex} shows example sentences from the dataset. The first tweet is a positive comment about the target and the sentence is also positive overall. The second tweet indicates a negative opinion about the target, since it has stated as crashed, although the sentence expresses a positive situation overall. Both sentence and targeted sentiment are the same for most of the tweets as in the first example. Only in 21\% of the tweets, targeted sentiment differs from the overall sentence sentiment. This means that the rest of the dataset is similar to a standard sentiment analysis dataset. The number of negative tweets in the dataset is significantly higher than the number of positive and neutral tweets for each target. The strikingly high number of negative tweets may be caused by the tendency of customers to write a review when they have had a bad experience. The total percentages of positive, negative and neutral classes are 19\%, 58\% and 23\%, respectively. The dataset is randomly divided into train, test, and validation sets by 65\%, 20\% and 15\%, respectively. The distribution of labels for each subset is kept similar to the distribution of labels for the entire dataset.
 

The dataset contains ungrammatical text, slang, acronyms, as well as special Twitter characters. During pre-processing  URLs and mentions (\textit{@}) are deleted. Hashtag signs (\textit{\#}) are removed, but hashtags are kept for two reasons: hashtags have been shown to express sentiment \cite{alfina2017utilizing, celebi2018segmenting} and some tweets contain the targets as hashtags.

\section{Methodology}

\citet{baldini-soares-etal-2019-matching} has introduced a novel method with transformer structure in the field of relation extraction. The key idea behind this work is to tag the entities with additional tokens before feeding the BERT model with the input. Different combinations of input and output types are evaluated. The best results are obtained when entity markers are added to the input and when the output of the starting entity markers are taken as the output from BERT. Motivated by the results of \citeauthor{baldini-soares-etal-2019-matching}'s work, this paper evaluates several BERT architectures with different input and output techniques for the targeted sentiment analysis task.

Two input representation techniques are investigated. In the standard input representation, the inputs are simply entered into the model without modification. In the second input representation approach, the targets are highlighted by adding additional special target tokens [TAR] at the beginnings and ends of targets, as shown in Table \ref{tbl:target_token}. These target tokens are expected to collect information about the target, just as the [CLS] token collects knowledge about the entire sentence. The three approaches for outputs explored in this study are shown in Figure \ref{fig:output}. The [CLS] output approach uses only the output of the first token from the last hidden state of BERT, as proposed for classification in the original paper \cite{devlin-etal-2019-bert}. In the second approach, the outputs of the tokens originating from the target, including the outputs of the [TAR] tokens, are  max-pooled. The first target marker approach considers only the output of the first [TAR] token in the input instead of the output of the standard [CLS]. All output approaches utilize a softmax layer at the end for classification.

\begin{table}
\centering
\begin{tabular}{ p{7.5cm} }
\hline
\textbf{Tweets with [TAR] tokens} \\
\hline

[TAR]\textit{whatsapp}[TAR] çöktü de biraz rahatladım bildirimlerden kurtuldum \newline ([TAR]\textit{whatsapp}[TAR] is crashed so I'm little relieved, got rid of notifications) \\

[TAR]\textit{coca cola}[TAR] daha iyi lezzet olarak \newline ([TAR]\textit{coca cola}[TAR]'s taste is better)  \\

\hline
\end{tabular}
\caption{
Example tweets with target marker representation
}
\label{tbl:target_token}
\end{table}

\subsection{Model Descriptions}

First, two baseline models are defined in order to show the drawbacks of the traditional SA models. One baseline is the BERT-based BounTi model \cite{koksal2021twitter}. The second baseline is also a BERT-based traditional SA model, but fine-tuned with our new dataset using sentence sentiment. Both have similar architectures and use the [CLS] output for sentiment classification. 

Four other variants of BERT-based models are proposed for targeted sentiment analysis. \textbf{T-BERT} is a model with a similar architecture to our baseline models. It makes no changes to the input and takes its output from the [CLS] token. The main difference is that targeted sentiment labels are used in the training phase. Therefore, the model is trained to learn targeted sentiment, whereas the baseline models are not aware of the target. \textbf{T-BERT\textsubscript{marked}} employs only the target marker representation on top of T-BERT and adds [TAR] tokens into the input. [TAR] token is introduced to BERT's tokenizer and the vocabulary is resized. Hence, the tokenizer accepts [TAR] as one of its special tokens such as [SEP]. \textbf{T-BERT\textsubscript{marked}-MP} is another model with target marker representation, additionally it max-pools all outputs of target tokens. \textbf{T-BERT\textsubscript{marked}-TS} also utilizes target markers. However, it takes its output only from the first target token [TAR] unlike T-BERT\textsubscript{marked}-MP.

 In the training phase of all models, BERTurk \cite{berturk} is chosen as the base BERT model. Class weights are set inversely proportional to the class distribution to reduce the effects of an unbalanced data set. The batch size is chosen as 24. Hyperparameters like weight decay, learning rate, and warm-up steps are selected as 0.1, $1e-5$, and $300$ respectively. As optimizer, AdamW is used.

\section{Results}

All proposed BERT variants and baselines are evaluated for targeted sentiment analysis over our introduced dataset. Macro averaged F1-Score is used as the evaluation metric in these experiments. The results are presented in Table \ref{tbl:tsa_result}. All targeted BERT variants outperform both baseline models for TSA. T-BERT\textsubscript{marked}-MP achieves the best results with  67\% F1-score, while T-BERT is relatively the worst performing targeted model with 61\% F1-score. T-BERT\textsubscript{marked}-TS and T-BERT\textsubscript{marked} obtain performance quite close to each other, the difference between those models is insignificant. They both have approximately 65\% F1-scores.

\begin{table}
\centering
\begin{tabular}{ p{5cm}  p{1.5cm} }
\hline
\textbf{Model} & \textbf{F1-Score} \\
\hline
Baseline Model  & 0.591 \\
\hline
BounTi Model  & 0.498\\
\hline
\hline
T-BERT  & 0.610 \\
\hline
T-BERT\textsubscript{marked}  & 0.659\\
\hline
T-BERT\textsubscript{marked}-TS  & 0.653\\
\hline
T-BERT\textsubscript{marked}-MP  & 0.669\\
\hline
\end{tabular}
\caption{
Performance of all models for TSA with test dataset against targeted sentiment labels
}
\label{tbl:tsa_result}
\end{table}

\begin{table}
\centering
\begin{tabular}{ p{5cm}  p{1.5cm} }
\hline
\textbf{Model} & \textbf{F1-Score} \\
\hline
Baseline Model & 0.256 \\
\hline
BounTi Model & 0.233\\
\hline
\hline
T-BERT & 0.401 \\
\hline
T-BERT\textsubscript{marked} & 0.428\\
\hline
T-BERT\textsubscript{marked}-TS & 0.459\\
\hline
T-BERT\textsubscript{marked}-MP & 0.444\\
\hline
\end{tabular}
\caption{
Performance of all models for TSA with data whose targeted and sentence sentiment are different.
}
\label{tbl:tsa_diff}
\end{table}

Only 21\% of the dataset has different sentence and targeted sentiment. These portion of data can demonstrate the distinction between targeted and sentence sentiment classification better. If both labels are the same, then traditional SA models may seem to accurately predict targeted sentiment. However, such sentences do not show how accurate the predictions from neither TSA nor SA models are. For this reason, a subset of our dataset such that all sentences have different targeted and sentence sentiment is used for another round of experiments. Table \ref{tbl:tsa_diff} shows the results for the TSA task with this subset. Baseline models' F1-score decreases dramatically to 25\%, and it's 23\% for BounTi model. Targeted BERT model with the lowest score (40\% F1-score) outperforms both models. T-BERT\textsubscript{marked}-TS achieves better targeted sentiment predictions with 46\% F1-score. T-BERT\textsubscript{marked}-TS improves the baseline performance by 79\% on F1-score. 

\section{Discussion}

Our results suggest that target oriented models can significantly improve the performance for targeted sentiment analysis. BERT architectures that perform successfully in the relation extraction field are shown to be successful for the targeted sentiment analysis task. Target markers make BERT models understand target related context better compared to the [CLS] token. All three models with target markers outperform the baselines and T-BERT. Hence, adding target markers is an effective approach for improving TSA performance. 

T-BERT\textsubscript{marked}-TS and T-BERT\textsubscript{marked}-MP are shown to perform slightly better than the other target oriented models. The common aspect of these models, apart from the target tokens, is that they both focus on the outputs of the target-related tokens rather than the [CLS] tokens. Therefore, it can be concluded that target outputs improves the performance for the TSA task.

We only considered one target in each sentence and annotated according to that target. Other targets in the sentence, if any, are ignored. Multiple targets with conflicting targeted sentiment in the same sentence can be a problem to consider. There are cases where a sentence has more than one target, and each target has a different targeted sentiment. For example, in a comparison, the sentiment toward one target may actually depend on the sentiment of another target in the same sentence. In this work, the scope is limited to only one target in each sentence. Target markers are also used only for this one target in the sentence and other possible targets are ignored. The lack of proper treatment of such cases in this work may affect the performance of all models.

Sentence and targeted sentiment are identical for 79\% of the dataset. 
Thus, if a traditional SA model, which is designed to predict the overall sentence sentiment, is used for the TSA task, its success for this task would be overestimated. 
The results demonstrate that targeted sentiment analysis models perform significantly better than traditional sentiment analysis models on the TSA task. However, the performance of the TSA models increases when they are tested on the entire test dataset, rather than on a subset containing only tweets with different sentence and targeted sentiment labels. This highlights that they may still be biased in favor of sentence sentiment to some extent.

\section{Ethical Considerations and Limitations}

The dataset contains public tweets in Turkish that are provided by the official Twitter API for research. Only tweet ID's and labels of the tweets are shared publicly to follow Twitter's terms and conditions. The annotators have no affiliation with any of the companies that are used as targets in the dataset, so there is no potential bias due to conflict of interest. 

The models developed in this work are not yet satisfactory to use their results without human monitoring. It is recommended to manually check the predictions of these models before using them.

\section{Conclusion and Future Work}

We presented a manually annotated Turkish Twitter dataset specifically created for targeted sentiment analysis and is also suitable for the traditional sentiment analysis task. This allowed us to develop and evaluate novel models for targeted sentiment analysis in a low-resource language such as Turkish.

We adapted and investigated BERT-based models with different architectures for targeted sentiment analysis. Experiments show significant improvement on baseline performance. 

As future work, we plan to expand our dataset so that it  contains more sentences with different sentence and targeted sentiment. Moreover, novel methods for sentences with multiple targets will be investigated.

\section*{Acknowledgements}
We would like to thank Abdullatif Köksal for helpful discussions and Merve Yılmaz Mutlu for annotations. GEBIP Award of the Turkish Academy of Sciences (to A.Ö.) is gratefully acknowledged.

\bibliography{anthology,custom}

\begin{thebibliography}{26}
\expandafter\ifx\csname natexlab\endcsname\relax\def\natexlab#1{#1}\fi

\bibitem[{Alfina et~al.(2017)Alfina, Sigmawaty, Nurhidayati, and
  Hidayanto}]{alfina2017utilizing}
Ika Alfina, Dinda Sigmawaty, Fitriasari Nurhidayati, and Achmad~Nizar
  Hidayanto. 2017.
\newblock Utilizing hashtags for sentiment analysis of tweets in the political
  domain.
\newblock In \emph{Proceedings of the 9th international conference on machine
  learning and computing}, pages 43--47.

\bibitem[{Baktha and Tripathy(2017)}]{baktha2017investigation}
Kiran Baktha and BK~Tripathy. 2017.
\newblock Investigation of recurrent neural networks in the field of sentiment
  analysis.
\newblock In \emph{2017 International Conference on Communication and Signal
  Processing (ICCSP)}, pages 2047--2050. IEEE.

\bibitem[{Baldini~Soares et~al.(2019)Baldini~Soares, FitzGerald, Ling, and
  Kwiatkowski}]{baldini-soares-etal-2019-matching}
Livio Baldini~Soares, Nicholas FitzGerald, Jeffrey Ling, and Tom Kwiatkowski.
  2019.
\newblock \href {https://doi.org/10.18653/v1/P19-1279} {Matching the blanks:
  Distributional similarity for relation learning}.
\newblock In \emph{Proceedings of the 57th Annual Meeting of the Association
  for Computational Linguistics}, pages 2895--2905, Florence, Italy.
  Association for Computational Linguistics.

\bibitem[{Birjali et~al.(2021)Birjali, Kasri, and Beni-Hssane}]{survey}
Marouane Birjali, Mohammed Kasri, and Abderrahim Beni-Hssane. 2021.
\newblock A comprehensive survey on sentiment analysis: Approaches, challenges
  and trends.
\newblock \emph{Knowledge-Based Systems}, page 107134.

\bibitem[{Celebi and {\"O}zg{\"u}r(2018)}]{celebi2018segmenting}
Arda Celebi and Arzucan {\"O}zg{\"u}r. 2018.
\newblock Segmenting hashtags and analyzing their grammatical structure.
\newblock \emph{Journal of the Association for Information Science and
  Technology}, 69(5):675--686.

\bibitem[{Cohen(1960)}]{cohen}
Jacob Cohen. 1960.
\newblock A coefficient of agreement for nominal scales.
\newblock \emph{Educational and psychological measurement}, 20(1):37--46.

\bibitem[{Dashtipour et~al.(2016)Dashtipour, Poria, Hussain, Cambria, Hawalah,
  Gelbukh, and Zhou}]{englishsa}
Kia Dashtipour, Soujanya Poria, Amir Hussain, Erik Cambria, Ahmad~YA Hawalah,
  Alexander Gelbukh, and Qiang Zhou. 2016.
\newblock Multilingual sentiment analysis: state of the art and independent
  comparison of techniques.
\newblock \emph{Cognitive computation}, 8(4):757--771.

\bibitem[{Devlin et~al.(2019)Devlin, Chang, Lee, and
  Toutanova}]{devlin-etal-2019-bert}
Jacob Devlin, Ming-Wei Chang, Kenton Lee, and Kristina Toutanova. 2019.
\newblock \href {https://doi.org/10.18653/v1/N19-1423} {{BERT}: Pre-training of
  deep bidirectional transformers for language understanding}.
\newblock In \emph{Proceedings of the 2019 Conference of the North {A}merican
  Chapter of the Association for Computational Linguistics: Human Language
  Technologies, Volume 1 (Long and Short Papers)}, pages 4171--4186,
  Minneapolis, Minnesota. Association for Computational Linguistics.

\bibitem[{Ding et~al.(2008)Ding, Liu, and Yu}]{ding2008holistic}
Xiaowen Ding, Bing Liu, and Philip~S Yu. 2008.
\newblock A holistic lexicon-based approach to opinion mining.
\newblock In \emph{Proceedings of the 2008 international conference on web
  search and data mining}, pages 231--240.

\bibitem[{Dong et~al.(2014)Dong, Wei, Tan, Tang, Zhou, and
  Xu}]{dong-etal-2014-adaptive}
Li~Dong, Furu Wei, Chuanqi Tan, Duyu Tang, Ming Zhou, and Ke~Xu. 2014.
\newblock \href {https://doi.org/10.3115/v1/P14-2009} {Adaptive recursive
  neural network for target-dependent {T}witter sentiment classification}.
\newblock In \emph{Proceedings of the 52nd Annual Meeting of the Association
  for Computational Linguistics (Volume 2: Short Papers)}, pages 49--54,
  Baltimore, Maryland. Association for Computational Linguistics.

\bibitem[{Jiang et~al.(2011)Jiang, Yu, Zhou, Liu, and
  Zhao}]{jiang-etal-2011-target}
Long Jiang, Mo~Yu, Ming Zhou, Xiaohua Liu, and Tiejun Zhao. 2011.
\newblock \href {https://aclanthology.org/P11-1016} {Target-dependent {T}witter
  sentiment classification}.
\newblock In \emph{Proceedings of the 49th Annual Meeting of the Association
  for Computational Linguistics: Human Language Technologies}, pages 151--160,
  Portland, Oregon, USA. Association for Computational Linguistics.

\bibitem[{Kaya et~al.(2012)Kaya, Fidan, and Toroslu}]{kaya}
Mesut Kaya, Guven Fidan, and Ismail Toroslu. 2012.
\newblock Sentiment analysis of turkish political news.
\newblock pages 174--180.

\bibitem[{K{\"o}ksal and {\"O}zg{\"u}r(2021)}]{koksal2021twitter}
Abdullatif K{\"o}ksal and Arzucan {\"O}zg{\"u}r. 2021.
\newblock Twitter dataset and evaluation of transformers for turkish sentiment
  analysis.
\newblock In \emph{2021 29th Signal Processing and Communications Applications
  Conference (SIU)}, pages 1--4. IEEE.

\bibitem[{Kruspe et~al.(2020)Kruspe, H{\"a}berle, Kuhn, and
  Zhu}]{kruspe-etal-2020-cross}
Anna Kruspe, Matthias H{\"a}berle, Iona Kuhn, and Xiao~Xiang Zhu. 2020.
\newblock \href {https://aclanthology.org/2020.nlpcovid19-acl.14}
  {Cross-language sentiment analysis of {European} {Twitter} messages during
  the {COVID-19} pandemic}.
\newblock In \emph{Proceedings of the 1st Workshop on {NLP} for {COVID-19} at
  {ACL} 2020}, Online. Association for Computational Linguistics.

\bibitem[{Li et~al.(2019)Li, Bing, Zhang, and Lam}]{li-etal-2019-exploiting}
Xin Li, Lidong Bing, Wenxuan Zhang, and Wai Lam. 2019.
\newblock \href {https://doi.org/10.18653/v1/D19-5505} {Exploiting {BERT} for
  end-to-end aspect-based sentiment analysis}.
\newblock In \emph{Proceedings of the 5th Workshop on Noisy User-generated Text
  (W-NUT 2019)}, pages 34--41, Hong Kong, China. Association for Computational
  Linguistics.

\bibitem[{Ma et~al.(2017)Ma, Li, Zhang, and Wang}]{ma2017interactive}
Dehong Ma, Sujian Li, Xiaodong Zhang, and Houfeng Wang. 2017.
\newblock Interactive attention networks for aspect-level sentiment
  classification.
\newblock In \emph{Proceedings of the 26th International Joint Conference on
  Artificial Intelligence}, pages 4068--4074.

\bibitem[{M{\"a}ntyl{\"a} et~al.(2018)M{\"a}ntyl{\"a}, Graziotin, and
  Kuutila}]{gov_org_comp}
Mika~V M{\"a}ntyl{\"a}, Daniel Graziotin, and Miikka Kuutila. 2018.
\newblock The evolution of sentiment analysis—a review of research topics,
  venues, and top cited papers.
\newblock \emph{Computer Science Review}, 27:16--32.

\bibitem[{Nguyen and Shirai(2015)}]{nguyen-shirai-2015-phrasernn}
Thien~Hai Nguyen and Kiyoaki Shirai. 2015.
\newblock \href {https://doi.org/10.18653/v1/D15-1298} {{P}hrase{RNN}: Phrase
  recursive neural network for aspect-based sentiment analysis}.
\newblock In \emph{Proceedings of the 2015 Conference on Empirical Methods in
  Natural Language Processing}, pages 2509--2514, Lisbon, Portugal. Association
  for Computational Linguistics.

\bibitem[{Pei et~al.(2019)Pei, Sun, and Li}]{targetedsentimentanalysis}
Jiaxin Pei, Aixin Sun, and C.~Li. 2019.
\newblock Targeted sentiment analysis: A data-driven categorization.
\newblock \emph{ArXiv}, abs/1905.03423.

\bibitem[{Schweter(2020)}]{berturk}
Stefan Schweter. 2020.
\newblock \href {https://doi.org/10.5281/zenodo.3770924} {Berturk - bert models
  for turkish}.

\bibitem[{Severyn and Moschitti(2015)}]{severyn-moschitti-2015-unitn}
Aliaksei Severyn and Alessandro Moschitti. 2015.
\newblock \href {https://doi.org/10.18653/v1/S15-2079} {{UNITN}: Training deep
  convolutional neural network for {T}witter sentiment classification}.
\newblock In \emph{Proceedings of the 9th International Workshop on Semantic
  Evaluation ({S}em{E}val 2015)}, pages 464--469, Denver, Colorado. Association
  for Computational Linguistics.

\bibitem[{Sun et~al.(2019)Sun, Huang, and Qiu}]{sun-etal-2019-utilizing}
Chi Sun, Luyao Huang, and Xipeng Qiu. 2019.
\newblock \href {https://doi.org/10.18653/v1/N19-1035} {Utilizing {BERT} for
  aspect-based sentiment analysis via constructing auxiliary sentence}.
\newblock In \emph{Proceedings of the 2019 Conference of the North {A}merican
  Chapter of the Association for Computational Linguistics: Human Language
  Technologies, Volume 1 (Long and Short Papers)}, pages 380--385, Minneapolis,
  Minnesota. Association for Computational Linguistics.

\bibitem[{Tang et~al.(2016)Tang, Qin, Feng, and Liu}]{tang2016effective}
Duyu Tang, Bing Qin, Xiaocheng Feng, and Ting Liu. 2016.
\newblock Effective lstms for target-dependent sentiment classification.
\newblock In \emph{Proceedings of COLING 2016, the 26th International
  Conference on Computational Linguistics: Technical Papers}, pages 3298--3307.

\bibitem[{Vaswani et~al.(2017)Vaswani, Shazeer, Parmar, Uszkoreit, Jones,
  Gomez, Kaiser, and Polosukhin}]{vaswani2017attention}
Ashish Vaswani, Noam Shazeer, Niki Parmar, Jakob Uszkoreit, Llion Jones,
  Aidan~N Gomez, {\L}ukasz Kaiser, and Illia Polosukhin. 2017.
\newblock Attention is all you need.
\newblock \emph{Advances in neural information processing systems}, 30.

\bibitem[{Wang et~al.(2016)Wang, Huang, Zhu, and Zhao}]{wang2016attention}
Yequan Wang, Minlie Huang, Xiaoyan Zhu, and Li~Zhao. 2016.
\newblock Attention-based lstm for aspect-level sentiment classification.
\newblock In \emph{Proceedings of the 2016 conference on empirical methods in
  natural language processing}, pages 606--615.

\bibitem[{Zhang et~al.(2020)Zhang, Chen, Liu, He, and
  Leung}]{zhang-etal-2020-target}
Ji~Zhang, Chengyao Chen, Pengfei Liu, Chao He, and Cane Wing-Ki Leung. 2020.
\newblock \href {https://doi.org/10.1162/tacl_a_00308} {Target-guided
  structured attention network for target-dependent sentiment analysis}.
\newblock \emph{Transactions of the Association for Computational Linguistics},
  8:172--182.

\end{thebibliography}
\bibliographystyle{acl_natbib}

\end{document}